\documentclass[journal]{IEEEtran}
\usepackage[pdftex]{graphicx}
\ifCLASSINFOpdf
\else
\fi
\usepackage{amsmath}
\usepackage{array}

\ifCLASSOPTIONcompsoc
 \usepackage[caption=false,font=normalsize,labelfont=sf,textfont=sf]{subfig}
\else
 \usepackage[caption=false,font=footnotesize]{subfig}
\fi
\usepackage{diagbox}
\usepackage{multirow}

\begin{document}
%

\title{Is Gender ``In-the-Wild'' Inference\\ Really a Solved Problem?}
%
%
%

\author{Tiago~Roxo 
        and Hugo~Proença,~\IEEEmembership{Senior Member, ~IEEE}
\thanks{Tiago Roxo is with the University of Beira Interior, Portugal, E-mail:tiago.roxo@ubi.pt.}
\thanks{Hugo Proença is with the IT: Instituto  de Telecomunicações, Department of  Computer Science, University of Beira Interior, Portugal, E-mail:hugomcp@di.ubi.pt.}
}

%
%

\markboth{arXiv, 2021}%
{Shell \MakeLowercase{\textit{et al.}}: Bare Demo of IEEEtran.cls for IEEE Journals}

%



\maketitle


\begin{abstract}

Soft biometrics analysis is seen as an important research topic, given its relevance to various applications. However, even though it is frequently seen as a solved task, it can still be very hard to perform in wild conditions, under varying image conditions, uncooperative poses, and occlusions. Considering the \textit{gender} trait as our topic of study, we report an extensive analysis of the feasibility of its inference regarding image (resolution, luminosity, and blurriness) and subject-based features (face and body keypoints confidence). Using three state-of-the-art datasets (PETA, PA-100K, RAP) and five Person Attribute Recognition models, we correlate feature analysis with gender inference accuracy using the Shapley value, enabling us to perceive the importance of each image/subject-based feature. Furthermore, we analyze face-based gender inference and assess the pose effect on it. Our results suggest that: 1) image-based features are more influential for low-quality data;  2) an increase in image quality translates into higher subject-based feature importance; 3) face-based gender inference accuracy correlates with image quality increase; and 4) subjects' frontal pose promotes an implicit attention towards the face. The reported results are seen as a basis for subsequent developments of inference approaches in uncontrolled outdoor environments, which typically correspond to visual surveillance conditions.
\end{abstract}



%
\IEEEpeerreviewmaketitle

\section{Introduction}
%
%
%
%

\begin{figure*}[!htb]
\centering
\includegraphics[width=7in]{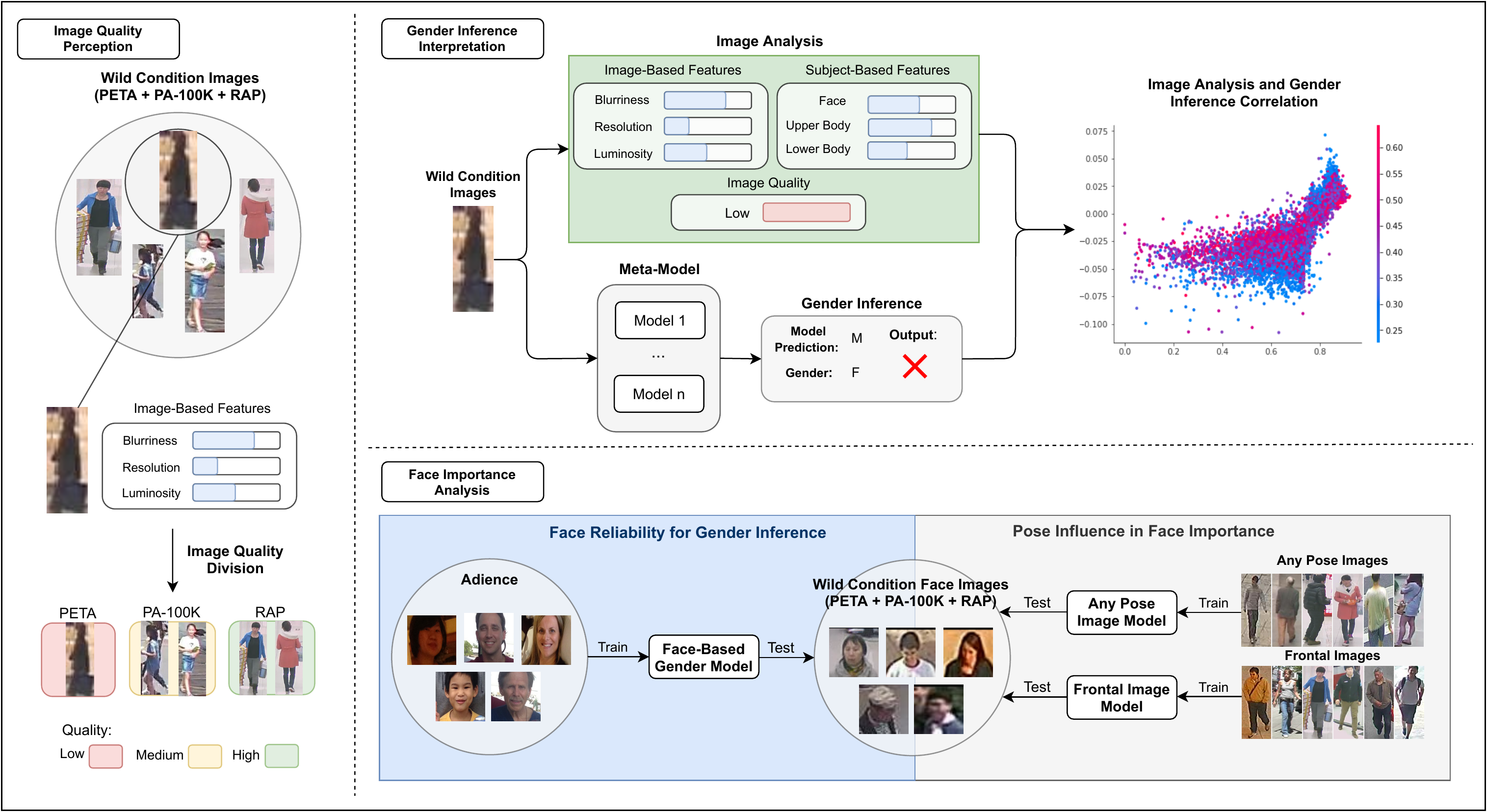}
\caption{
%
Overview of the used approaches for image and subject-based features correlation with gender inference. We divide the work into three major phases: 1) image quality perception; 2) gender inference interpretation; and 3) face importance analysis. We divide the PAR datasets considering image-based features and correlate image and subject-based features to gender inference accuracy of a meta-model. This model is a combination of five PAR models. The face importance is analyzed in two ways: by evaluating its reliability in wild conditions and assessing pose influence in face-based gender inference.
}
\label{fig:main_image}
\end{figure*}


\IEEEPARstart{S}{oft} biometrics analysis has been gaining increased interest over the last years, given its applicability to aid in person identification while requiring little to no cooperation from observed subjects and being robust to low quality data \cite{jain2004soft, dantcheva2015else}. This interest is also enhanced by the availability of massive amounts of data in wild conditions, provided by surveillance cameras and footage gathered by hand-held devices.  

Even though soft biometrics inference is frequently seen as a relatively easy task, this work reports results that point to the opposite.  In particular, we focus our extensive analysis on the \textit{gender} trait, posing the following research questions:

\begin{itemize}
    \item \textbf{How well solved is \textit{in-the-wild} gender inference?}
    \item  \textbf{What features have the most relevance for gender inference accuracy?}
    \item \textbf{Given the face importance in gender inference, what conditions justify its use?}
\end{itemize}

Gender information can be retrieved from an image containing the human silhouette and face. As such, the conclusions yielding from the answers to the above questions might apply to other soft biometrics traits with similar characteristics, such as \textit{age}, \textit{body mass}, or \textit{ethnicity}.

Gender inference \textit{in-the-wild}, such as in surveillance environments, faces various challenges (e.g., occlusions, subjects' pose, image resolution, and lighting conditions) that strongly decrease the state-of-the-art deep learning-based models' accuracy. When comparing the conditions between cooperative and \textit{in-the-wild} environments, subjects' pose and image quality are key discriminating factors that might influence inference accuracy. As such, in this work, we correlate image and subject-based features' importance with models' accuracy to perceive the importance of each feature in gender inference. To achieve this goal, we define quantifiable image and subject-based features to categorize the datasets' quality. We define \emph{resolution}, \textit{luminosity}, and \textit{blurriness} as the most influential image-based features and the \textit{keypoints} (KP) \textit{confidence} of the \textit{face}, \textit{upper}, and \textit{lower body} parts as the subject-based features. The KP data is obtained from Alphapose \cite{fang2017rmpe, li2018crowdpose, xiu2018poseflow}, a state-of-the-art pose estimator.

Given the intent to evaluate gender inference in wild conditions, we use
the most challenging Pedestrian Attribute Recognition (PAR) datasets in our experiments. Furthermore, to obtain a generic deep learning model performance evaluation, we gather five state-of-the-art PAR inference models and combine their predictions to obtain a model-independent gender inference. To objectively perceive features' importance, we correlate them with the models' combined inference and analyze it using the Shapley value \cite{shapley1953value}, a game-theoretic approach to explain machine learning outputs.


Another important analysis in our work regards the usability of the facial region for gender inference. Given that the face is intuitively one of the most important attributes for gender inference, several models have been proposed using only the face for this task. However, the availability of the facial region in wild situations is not a guaranteed condition, and a subject's pose might influence its importance for face-based gender inference. To evaluate these concepts, we start by creating challenging (wild) face sets, yielding from frontal images of various PAR datasets; frontal images are obtained from a pose-based image division, using Alphapose. Then, we compare a face-based model's performance trained in Adience - a face-based gender dataset - and evaluate it on the created face sets. Furthermore, we assess the effect of pose in face-based gender inference accuracy by training the model in \textit{any} or \textit{frontal} pose full-body images and evaluating its performance in face datasets.

According to the above points, the main contributions of this work can be summarized as follows:

\begin{itemize}
    \item We define the most relevant image and subject-based features for gender inference and relate their importance with image quality;
    \item We describe an approach to obtain a novel wild face dataset from various PAR sources, using a pose estimation method for head detection and Region of Interest (ROI) definition;
    \item We evaluate the importance of facial regions for \textit{in-the-wild} gender inference, relating it with image quality;
    \item We report an implicit face attention mechanism for gender inference, via frontal pose image training.
\end{itemize}

The remainder of this paper is organized as follows: Section \ref{sec:gender-estimation} summarizes the most relevant face and body-based methods used for gender inference; Section \ref{sec:methodology} describes the methodologies used for our analysis and Section \ref{sec:experiments} discusses the results obtained. The main conclusions and future work are presented in Section \ref{sec:conclusion}.

\section{Gender Inference}
\label{sec:gender-estimation}

\subsection{Face-Based Approaches}


The facial region is commonly used for gender inference, with several datasets and models designed for this task. Levi and Hassner \cite{levi2015age} proposed a simple Convolutional Neural Network (CNN) model, with only five layers, given the nature of age and gender inference (small number of classes). Zhan \emph{et al.} \cite{zhang2017age}, used an approach based on Residual Networks of Residual Networks (RoR), which outperformed other CNN architectures. In the same context, Ranjan  \emph{et al.} \cite{ranjan2017hyperface} developed HyperFace, a multi-task framework based on a trunked Alexnet \cite{krizhevsky2012imagenet} for face detection, fused with a CNN for face landmarks detection and gender recognition.

The facial region usefulness in gender inference has been explored using real-time prediction \cite{arriaga2017real}, minimizing the number of parameters by combining \textit{Conv2D} and \textit{BatchNorm} blocks and avoiding \textit{Fully Connected} (FC) layers. Furthermore, Lee \emph{et al.} \cite{lee2018joint} described a lightweight inference approach that uses depthwise separable convolution layers to reduce the model size and save inference time.

The inference of facial attributes has also been reported in the literature. Liu \emph{et al.} \cite{liu2015deep} combined a Support Vector Machine (SVM) for face attribute inference with two CNNs: LNET for appropriately resized face localization and ANET for distinguishing face identities, using global and local convolutions for feature extraction. Ryu \emph{et al.} presented InclusiveFaceNet \cite{ryu2017inclusivefacenet}, proposing the inclusion of race and gender for face attribute detection, based on modifications of FaceNet \cite{schroff2015facenet}.

Regarding gender inference based on facial data, several datasets have been announced \cite{liu2015deep, rothe2018deep, cao2018vggface2, escalera2016chalearn}. All these datasets represent relatively cooperative (controlled) environments, with good image quality, being Adience \cite{eidinger2014age} the most challenging one.

\subsection{Body-Based Methods}

Given the conditions in surveillance scenarios, it is important to consider that the facial region might not be available, which supports the use of body-based models for gender inference. In this context, PAR is a topic broadly discussed in recent works, aiming to identify pedestrian attributes such as clothing, accessory usage, gender, and age, under challenging covariates such as occlusion, pose, image resolution, and luminosity. 

 Wang \emph{et al.} \cite{wang2017attribute} proposed a model based on a Recurrent Neural Network (RNN) encoder-decoder framework, using Long Short-Term Memory (LSTM) as a recurrent neuron. They promoted joint recurrent learning by using their framework in images divided into six horizontal strips. Zhao \emph{et al.} \cite{zhao2018grouping} also used a similar partition strategy to perform attribute recognition, combining LSTM and BN-Inception \cite{ioffe2015batch} networks. 

Attention-based approaches are also highly popular: Sarafianos \emph{et al.} \cite{sarafianos2018deep} combined primary and attention classifiers outputs to learn effective attention maps at multiple scales. Tang \emph{et al.} \cite{tang2019improving} focused on a feature pyramid structure-based model, with four different feature levels, each with Attribute Localization Modules (ALM). ALM is inspired by Spatial Transformer Network (STN) \cite{jaderberg2015spatial}, which discovers the discriminative regions for each attribute in a weakly supervised manner. Guo \emph{et al.} \cite{guo2019visual} defined a new attention consistency loss, translated by the distance between transformed attention heatmaps of original and flipped images. This approach provides more focused heatmaps, resulting in classification performance improvements.

The analysis of attributes' importance for pedestrian classification is also reported in the literature. Li \emph{et al.} \cite{li2015multi} performed global image analysis, exploring how one attribute contributes to the representation of others. They introduced DeepMAR, a learning framework that recognizes multiple attributes simultaneously, exploring the relationships among them. Lin \emph{et al.} \cite{lin2019improving} developed a re-identification model that considers the attribute importance for this goal. They introduced an Attribute Re-weighting Module (ARM), which corrects attribute predictions based on the learned dependency and 
correlation between them. Jia \emph{et al.} \cite{jia2020rethinking} showed that enhancing localization of attribute-specific areas, typically adopted by state-of-the-art methods, may not be beneficial for performance improvement.

For additional information on this topic, Wang \emph{et al.} \cite{wang2019pedestrian} review the main divisions of existing PAR approaches, providing a brief summary of the state-of-the-art PAR methods.

\section{Methodologies}
\label{sec:methodology}

An overview of our processing chain to analyze the effectiveness of gender inference and its correlation with different image features is shown in Fig. \ref{fig:main_image}. We divide three well known PAR datasets (PETA, PA-100K, and RAP) into three image quality levels, based on image-based feature analysis. Then, we correlate image and subject-based features with the gender inference accuracy of a meta-model, in order to perceive the image/subject-based features that favor/penalize overall accuracy. The used meta-model is a combination of the output of five different state-of-the-art PAR models.  Furthermore, we analyze the  importance of the facial region in gender inference, evaluating a face-based model performance in wild conditions and examining the importance of pose in face-based gender inference.






\subsection{Methods and Datasets}

For gender inference analysis, we use five models with publicly available implementations. We consider DeepMar \cite{li2015multi} and StrongBase \cite{jia2020rethinking}, which aim to infer pedestrian attributes based on global image analysis, and VAC \cite{guo2019visual} and ALM \cite{tang2019improving}, which consider attention mechanisms for this task. Additionally, we consider the PAR component of the re-identification method ARP \cite{lin2019improving}. Regarding the models backbones, all methods are based on ResNet50 \cite{he2016deep}, except for ALM, which is based on BN-Inception \cite{ioffe2015batch}. 

Regarding the selection of datasets, our goal was to use a set of images with a wide variability range to reproduce, as much as possible, \emph{in-the-wild} conditions. For this reason, we use PETA \cite{deng2014pedestrian}, PA-100K \cite{liu2017hydraplus}, and RAP \cite{li2016richly}, three state-of-the-art PAR datasets. The chosen datasets clearly display different levels of quality, as discussed in our experiments.


\subsection{In-the-Wild Major Variability Factors}

\begin{figure}[!tb]
\includegraphics[scale=0.45]{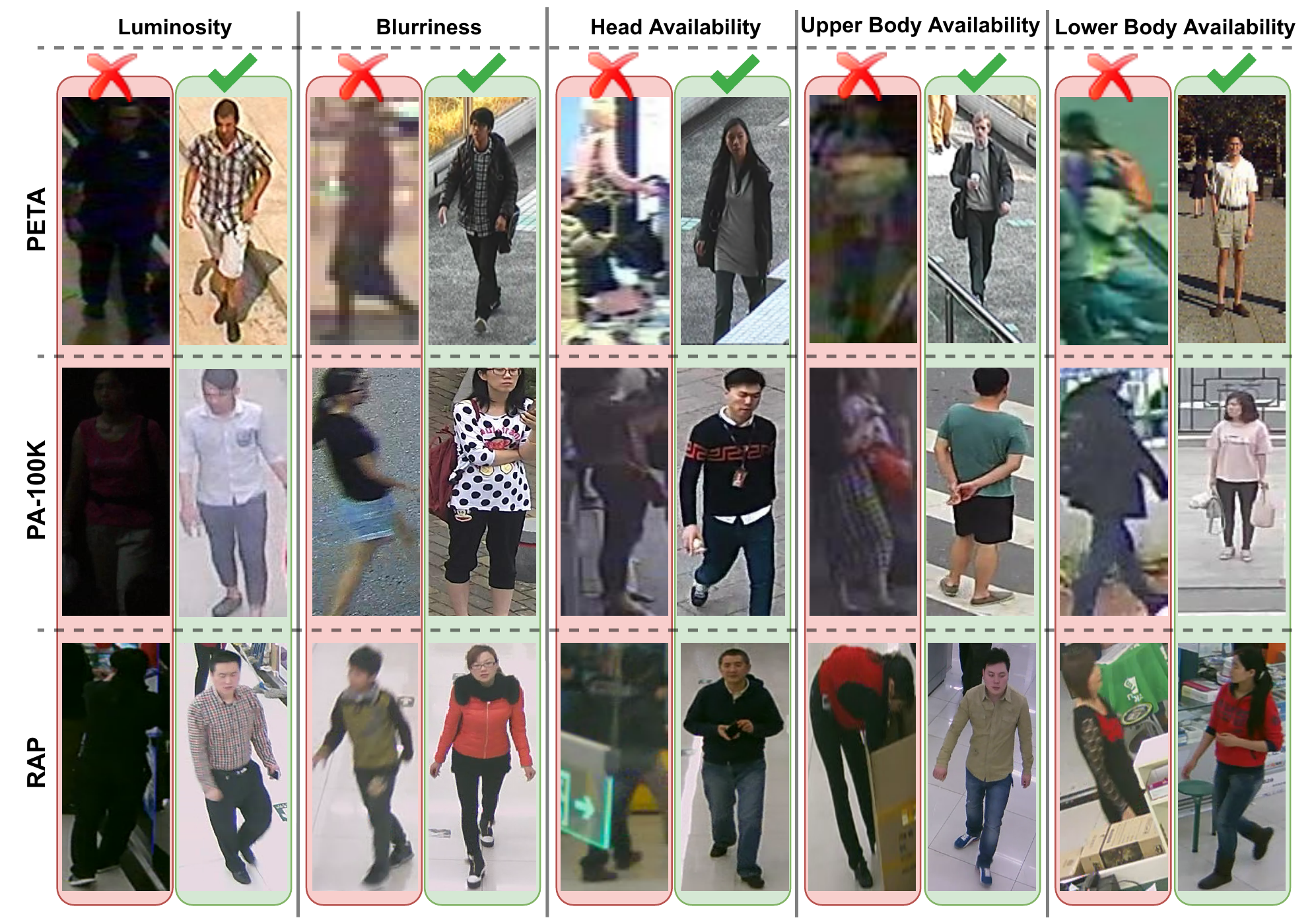}
\centering
\caption{Dataset examples of the evaluated features, divided into image (\textit{luminosity} and \textit{blurriness}) and subject-based (\textit{head}, \textit{upper body}, and \textit{lower body} availability) features. Each row corresponds to one of the three PAR datasets. Each column contains opposite cases of each feature, with the left one representing \emph{bad quality} (with feature value closer to its minimum) and the right one representing \emph{good quality}. For the blurriness feature, high feature value represents \emph{bad quality}. All images are resized to the same resolution for visualization purposes.}
\label{fig:dataset_examples}
\end{figure}

Our primary goal was to perceive what is a \emph{good quality} image, in wild conditions, for gender inference. This requires measuring the factors that typically degrade gender inference. We propose a division in image and subject-based features, described in the following subsections. Examples of image and subject-based features analyzed are shown in Fig. \ref{fig:dataset_examples}, displaying contrasting examples of each feature.

\subsubsection{Image-Based Features}

We consider \textit{blurriness}, \textit{luminosity}, and \textit{resolution} as the major image-based factors for gender inference. We retrieve image resolution by multiplying its width and height. For luminosity, we obtain the values of red, green, and blue channels to measure the perceived brightness \cite{lumino}. Blurriness yields from the convolution of the images with a Laplacian kernel, taking the variance as result. 


\subsubsection{Subject-Based Features}


We use Alphapose \cite{fang2017rmpe, li2018crowdpose, xiu2018poseflow}, an accurate multi-subject pose estimator, to extract subject-based features. Given that the used datasets contain subjects' ROI, we change the Alphapose implementation to ignore its You Only Look Once (YOLO) \cite{redmon2016you} detector input. This change translates into a more coherent pose estimation. In exceptional cases, where the image was not trimmed to the subject, we keep the original implementation. The Alphapose outputs 51 KP for each image, corresponding to the image coordinates $x$ and $y$, and the confidence score for each of the 17 Common Objects in Context (COCO) \cite{lin2014microsoft} KP: nose, left and right eye, left and right ear, left and right shoulder, left and right elbow, left and right wrist, left and right hip, left and right knee, and left and right ankle. We group the KP into three regions: \textit{face}, \textit{upper}, and \textit{lower body}. \textit{Face} confidence is the mean of nose, eyes, and ears KP confidences. \textit{Upper body} confidence is based on shoulders, elbows, and wrists, while \textit{lower body} confidence uses the values of hips, knees, and ankles.

\begin{figure}[!tb]
\centering
\includegraphics[scale=0.5]{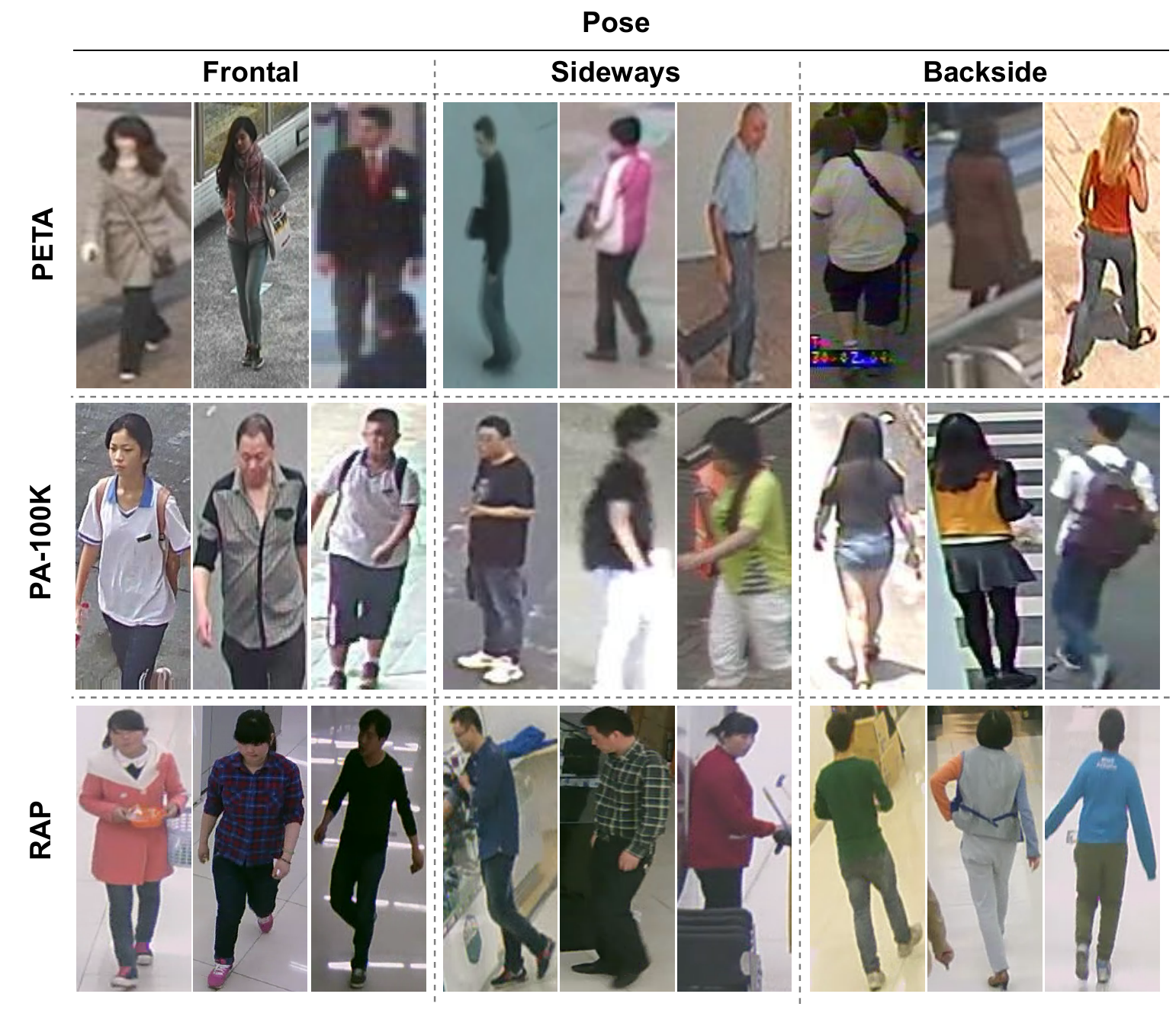}
\caption{Examples of images considered as \textit{frontal}, \textit{sideways} and \textit{backside}. Each row is related to one of the three PAR datasets and each column regards examples of the respective pose.}
\label{fig:people_orientation}
\end{figure}

Regarding subject poses, we consider three values: \textit{frontal}, \textit{sideways}, and \textit{backside}. We differentiate a subject facing forward or backward using pose estimation information. If the rightmost shoulder, from the top left of the image, is the left shoulder, the subject faces forward; otherwise, the subject is facing backward. A subject is considered \emph{sideways} if the shoulder length with respect to the upper body height is less than 0.5. Shoulder length is the absolute difference between the right and left shoulders, and the upper body height is the difference between the shoulder and hip. Some examples of images considered \textit{frontal}, \textit{sideways}, and \textit{backside} are presented in Fig. \ref{fig:people_orientation}.

\subsection{Shapley Value}
\label{method:shapley}

The Shapley value, coined by Shapley in 1953 \cite{shapley1953value}, is a cooperative game theory-based method used for assigning payouts to players, depending on their contribution towards the total payout. In the machine-learning context, the Shapley value is used to evaluate how each feature (player) of a given instance contributed (assigning payout) towards the model prediction of the instance (total payout). 





The use of Shapley values in our experiments is linked to our interest in analyzing how each feature contributed to gender inference and if it contributed positively or negatively. The higher the absolute Shapley value is, the higher the feature influence. We associated a value of 1 to instances where all models correctly inferred the gender and 0 otherwise. As such, positive values represent cases that influence correct gender inference, while negative values contribute to gender misinference. 
We correlate the proposed feature values with gender inference using the SHapley Additive exPlanations (SHAP) framework \cite{lundberg2017unified} and the \emph{TreeExplainer} model \cite{lundberg2020local}, from a publicly available implementation \cite{git_shap}.

\subsection{Head Detection} 



\begin{figure}[!tb]
\centering
\includegraphics[scale=0.7]{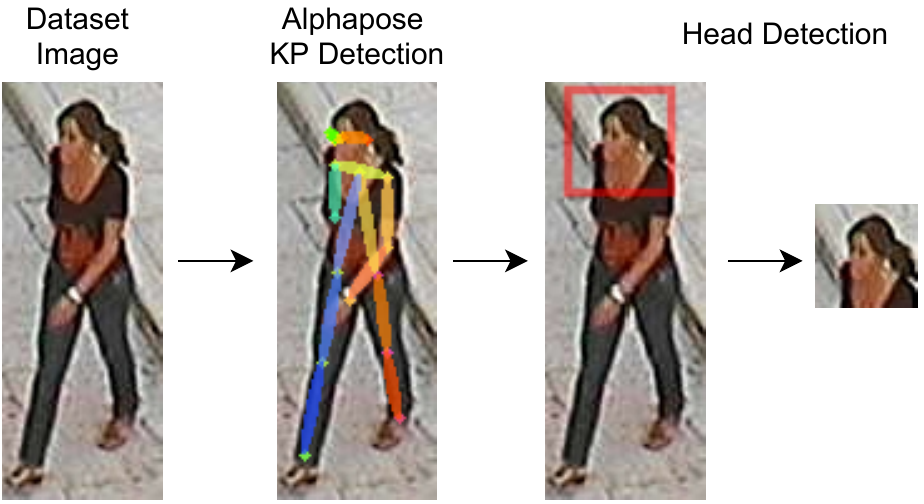}
\caption{Phases of the head detection process, displaying pose estimation and head bounding box size. Faces are obtained exclusively from the frontal images of each dataset.}
\label{fig:face_detection}
\end{figure}

Given the varying image quality, we use pose estimation to aid in head detection. Based on the proposed subjects' pose division, we crop the head regions from frontal images. First, we obtain the coordinates $x$ and $y$ of the right and left ears and assume that the head's central point is given by the arithmetic mean of such coordinates. Then, head ROIs are drawn using the top-left and bottom-right bounding boxes coordinates, centered in the previously calculated head's central point, always considering the head ROI height as $\frac{2}{9}$ of the whole body silhouette height. Head bounding boxes drawing is based on Detectron~\cite{detectron2}. We provide the steps involved in head detection in Fig. \ref{fig:face_detection}, illustrating pose estimation and head ROI definition.

\subsection{Face Importance Analysis}



We evaluate face importance in gender inference by training a face-based gender model in full-body images and obtaining the mean Accuracy ($mA$) \cite{li2016richly} in faces of frontal images of the corresponding dataset. The $mA$ criterion is the mean classification accuracy of the positive and negative samples for each attribute, averaged over all attributes:

\begin{equation}
mA = \dfrac{1}{2N} \sum_{i=1}^{M} \left(\dfrac{TP_i}{P_i} + \dfrac{TN_i}{N_i} \right),
\end{equation}

where \textit{N} is the number of examples, \textit{M} the number of attributes, \textit{$P_i$} is the number of positive examples and \textit{$TP_i$} is the number of correctly predicted positive examples for the $i^{th}$ attribute; \textit{$TN_i$} and \textit{$N_i$} are defined analogously.

To obtain an objective measure of face importance ($FI$), we transform the $mA$ values of frontal image faces ($mA_f$) as follows: 

\begin{equation}
    FI = \dfrac{mA_f - 50}{mA_{max} - 50}, 
\end{equation}

where the value 50 represents the model inference randomness, and $mA_{max}$ is the $mA$ value obtained when the model is evaluated in full-body images. 
FI value is enclosed in the $[0, 100]$ interval, directly corresponding to the face importance in each evaluated context.

\section{Experiments}
\label{sec:experiments}

\subsection{Dataset and Evaluation Metrics}


Our experiments are carried out in three state-of-the-art PAR datasets: PETA \cite{deng2014pedestrian}, PA-100K \cite{liu2017hydraplus}, and RAP \cite{li2016richly}.  The PETA dataset has 19,000 images, each with 61 binary attributes. It is divided into three subsets: 9,500 for training, 1,900 for validation, and 7,600 for testing. Based on this dataset's protocol, we report the results on the 35 attributes with a ratio of positive labels higher than 5\%. The PA-100K dataset is composed of 100,000 images from outdoor surveillance cameras. It is split into 80,000 images for training, 10,000 for validation, and 10,000 for testing. Each image has 26 attributes. The RAP dataset contains 41,585 images, each annotated with 72 attributes. Following the official protocol \cite{li2016richly}, we split the dataset into 33,268 training images and 8,317 test images and report the results on the 51 binary attributes with a positive ratio higher than 1\%.


For face-based gender inference, we use Adience~\cite{eidinger2014age}, a dataset composed of face images obtained from cameras, smartphones, and tablets. This dataset contains varying quality factors such as facial expressions, blurred samples, partial occlusions, and head pose variations. It is composed of 26,580 unconstrained images of 2,284 subjects. We use the original in-plane aligned version of the faces~\cite{eidinger2014age}.





To present our results, we use the label-based metric $mA$~\cite{li2016richly}. For state-of-the-art comparison purposes, we use accuracy when evaluating our face-based gender model (described in Section \ref{face-gender-model}) in the Adience dataset.

\subsection{Implementation Details}
\subsubsection{Baseline Methods}

The models used were VAC \cite{guo2019visual}, ARP \cite{lin2019improving}, ALM \cite{tang2019improving}, StrongBase \cite{jia2020rethinking}, and DeepMar \cite{li2015multi}. 


The VAC model train/test transformation for PETA and RAP is based on the WIDER dataset \cite{li2016human} transformation.  Since the attribute weights for weighted sigmoid cross-entropy \cite{li2015multi} loss were only defined for PA-100K, we retrieve the attribute weights for PETA and RAP from the same source the authors used to obtain the PA-100K ones. To allow gender inference comparison in our experiments, we change the PETA dataset evaluation for the StrongBase evaluation implementation, adjusting the output-related loss without modifying attention-related ones.

For ARP, we only consider attribute loss, ignoring the re-identification one. As the available implementation did not consider PETA, RAP, and PA-100K dataset evaluation, we introduce it following the original paper's guidelines.


The remaining models are evaluated using the original implementations. All the gender models used in our experiments are specifically trained for gender attribute only. For all models' implementations that did not consider the evaluated datasets, we incorporate StrongBase dataset reading into their implementation. For consistency purposes, all metric evaluations are also based on StrongBase's implementation.

\subsubsection{Face-Based Gender Model}
\label{face-gender-model}

The proposed model is based on the StrongBase implementation \cite{jia2020rethinking}. The ResNet50 \cite{he2016deep}, pretrained in ImageNet \cite{deng2009imagenet}, is used as backbone to extract image features. The model is separated into feature extraction and classification. The feature extractor is composed of ResNet50 without its \textit{Average Pool} and \textit{FC} layer, which were incorporated in the classifier. The classifier has its \textit{FC} layer replaced by a \textit{Linear} layer, followed by a \textit{Batch Normalization} one. Images are resized to $256\times192$ with random horizontal mirroring as inputs. Stochastic Gradient Descent (SGD) is used for training, with momentum of $0.9$, and weight decay of $0.0005$.  The initial learning rate is set to $0.01$ for the feature extractor and $0.1$ for the classifier. Plateau learning rate scheduler is used with a reduction factor of $0.1$ and a patience epoch number of $4$. Batch size is set to $64$ and the total epoch number of training is $30$, using Binary Cross Entropy with Logits (BCELogits) as loss function.

\subsection{Cross-Domain Performance}


We evaluate the gender inference performance of each model, when trained for all pedestrian attributes, in the \textit{within} and \textit{cross-domain} settings. This analysis allows us to perceive the potential robustness of each model. Additionally, we train each model specifically for the gender attribute and compare its performance with the corresponding model trained for all dataset attributes. We report our results in Table \ref{table:cross_domain}.

\begin{table}[!t]
    \centering
    \renewcommand{\arraystretch}{1.1}
    \caption{Gender $mA$ of different models trained and evaluated on various PAR datasets. Models named with \textit{Gender} suffix refer to models trained solely for the gender attribute. The outperforming methods for each dataset are shown in bold.}
    \begin{tabular}{|c|*{4}{c|}}\hline
        \backslashbox[20mm]{\textbf{Train}}{\textbf{Eval}} & \textbf{Methods} & \makebox[3em]{\textbf{PETA}}  & \makebox[3em]{\textbf{PA-100K}} & \makebox[3em]{\textbf{RAP}} \\\hline\hline
        
        \multirow{10}{*}{\textbf{PETA}} &
        StrongBase &
        92.80 & 
        78.98 &
        79.47 \\

         &
        StrongBase-Gender &
        \textbf{93.13} & 
        79.65 &
        80.44 \\
        
        \cline{2-5}
        
        &
        ALM &
        91.36 & 
        76.95 &
        80.62 \\
        
         &
        ALM-Gender & 
        92.28 & 
        78.27 &
        82.66 \\
        
        \cline{2-5}
        
        &
        DeepMar &
        91.08 & 
        72.83 &
        71.76 \\
        
        &
        DeepMar-Gender &
        92.33 & 
        77.63 &
        74.78 \\
        
        \cline{2-5}
        
        &
        VAC &
        92.85 & 
        78.16 &
        78.50 \\ 
        
        &
        VAC-Gender &
        92.76 & 
        78.22 &
        77.92 \\

        \cline{2-5}

        &
        APR &
        92.84 & 
        78.83 &
        78.95 \\
        
        &
        APR-Gender &
        92.34 & 
        75.49 &
        78.49 \\

        \hline
        \hline
        \multirow{10}{*}{\textbf{PA-100K}} &
        StrongBase &
        77.28  &
        90.97 &
        86.93 \\
        
        &
        StrongBase-Gender &
        75.01 &
        90.77 &
        88.38 \\

        \cline{2-5}
        
        &
        ALM  &
        75.48 &
        88.34 &
        85.87 \\
        
        &
        ALM-Gender &
        76.59 &
        90.34 &
        88.55 \\
        
        \cline{2-5}
        &
        DeepMar &
        76.17 &
        88.52 &
        85.73 \\
        
        &
        DeepMar-Gender &
        74.50 &
        90.39 &
        85.72 \\

        \cline{2-5}
        &
        VAC &
        75.20 &
        90.17 &
        87.79 \\

        &
        VAC-Gender &
        76.85 &
        \textbf{91.05} &
        88.98 \\
        
        \cline{2-5}
        &
        APR &
        76.53 &
        89.91 &
        87.24 \\
        
        &
        APR-Gender &
        75.26 &
        90.05 &
        86.29 \\
        
        \hline
        \hline
        \multirow{10}{*}{\textbf{RAP}} &
        StrongBase &
        67.67 &
        72.25  &
        \textbf{96.74} \\
        
        &
        StrongBase-Gender &
        71.41 &
        76.97 &
        96.34 \\

        \cline{2-5}
        
        &
        ALM &
        76.81 &
        81.24 &
        95.69 \\
        
        &
        ALM-Gender &
        76.03 &
        80.28 &
        95.41 \\

        \cline{2-5}
        &
        DeepMar &
        69.29 &
        75.76 &
        95.52 \\

        &
        DeepMar-Gender &
        70.22 &
        76.89 &
        96.40 \\
        
        \cline{2-5}
        
        &
        VAC &
        66.11 &
        68.84 &
        96.52 \\

        &
        VAC-Gender &
        72.89 &
        76.67 &
        96.59 \\
        
        \cline{2-5}
        &
        APR &
        68.39 &
        75.41 &
        96.20 \\
        
        &
        APR-Gender &
        69.24 &
        72.93 &
        95.90 \\

        \hline
    \end{tabular}
    \label{table:cross_domain}
\end{table}

From this experiment, we can observe that models trained exclusively with the gender label tend to have slightly better gender inference accuracy than those trained for all attributes. This is confirmed in both \textit{within} and \textit{cross-domain} scenarios. Therefore, in all the remaining experiments, we resort to models trained exclusively for gender.

Regarding the obtained models' accuracies, we conclude that no model is significantly better than the others, regardless of the dataset where it was trained/evaluated. However, when examining model performance within and across datasets, some differences can be observed. First, models trained in PETA or PA-100K and evaluated in other datasets tend to perform well, achieving results close to those they present when evaluated in the same dataset they were trained on. The exception is RAP-trained models, where they perform considerably worse if evaluated in other datasets. Second, considering the \textit{within-domain} setting, RAP is the dataset where models perform the best, achieving the highest accuracy out of all the three evaluated datasets. Finally, models trained in PETA or PA-100K achieve better performances in RAP than in any other dataset, aside from the one they were trained on. PETA is the dataset where models, if trained on other datasets, perform the worst. These observations indicate that accurate gender inference is easier in RAP images and that PETA is the most challenging dataset for this task.

\subsection{Dataset Analysis}

To perceive how image-based features influence gender inference, we obtain the corresponding values for each feature/dataset. We present the normalized mean and standard deviation values in Table \ref{table:dataset_analysis}.

\begin{figure}[!tb]
\centering
\includegraphics[scale=0.55]{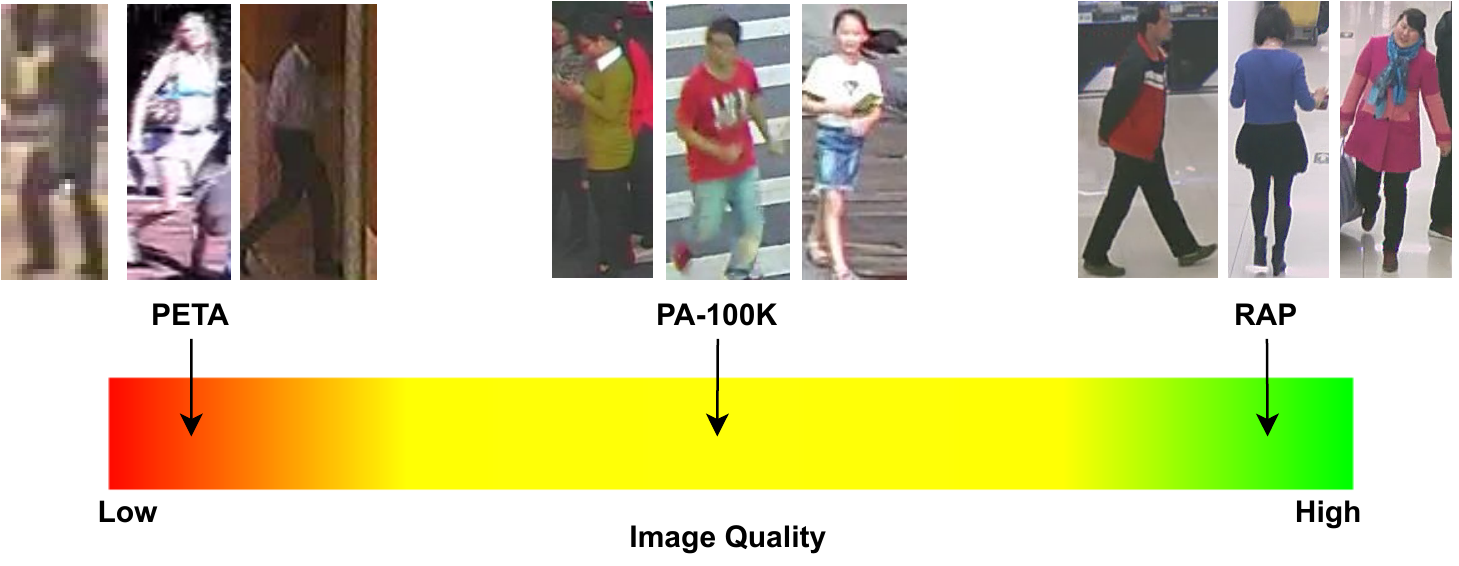}
\caption{
PAR dataset quality division, taking into account image-based feature values. Each dataset illustrates examples of the corresponding quality levels.
}
\label{fig:gradient_quality}
\end{figure}

Given the low variability of luminosity across the datasets, we group them into categories exclusively based on resolution and blurriness: \textit{low}, \textit{medium}, and \textit{high}. The higher quality dataset is RAP, given the high resolution value and sharpness. PETA is the dataset with lower quality, based on the low resolution values and high blurriness. Since PA-100K dataset feature values are typically between those of PETA and RAP, we consider this dataset as of medium quality. We present a representative quality bar of the three datasets in Fig. \ref{fig:gradient_quality}, illustrating some dataset examples.

\begin{table}[!t]
    \renewcommand{\arraystretch}{1.1}
    \centering
    \caption{Dataset image-based feature analysis. Values are normalized for the combination of the three dataset values, for each feature.}
    \begin{tabular}{|c|*{3}{c|}}\hline
        \textbf{Dataset}  & 
        \textbf{Resolution} &
        \textbf{Luminosity} &
        \textbf{Blurriness} 
        \\\hline\hline
        
        PETA &
        0.037 $\pm$ 0.024 &
        0.432 $\pm$ 0.100 &
        0.120 $\pm$ 0.148 \\

        PA-100K & 
        0.062 $\pm$ 0.063 &
        0.449 $\pm$ 0.126 & 
        0.095 $\pm$ 0.089 \\
        
        RAP & 
        0.131 $\pm$  0.083 &
        0.407 $\pm$  0.107 & 
        0.022 $\pm$  0.013 \\

 
        
        \hline
        
    \end{tabular}
    \label{table:dataset_analysis}
\end{table}

\subsection{Methods Evaluation}

We evaluate the image and subject-based features of the correctly inferred cases, in comparison to all test set images, to perceive the features that contribute the most to gender misinference. Here, we consider the \textit{within-domain} setting. Given the predictions of the five methods used, we consider \emph{correctly classified} cases exclusively when all models correctly infer the corresponding value.  Results are displayed in Table \ref{tabel:method_prediction}, with values normalized per dataset.

\begin{table*}[!t]
    \centering
    \caption{
    Image and subject-based feature values for PETA, PA-100K, and RAP datasets. \textit{Correct} represents the correctly inferred images (when all models inferred the correct gender value) and \textit{All} refers to all test data. Subjects pose values represent the portion of test images with the given pose.
    }
    \begin{tabular}{|cl|*{6}{c|}}\hline
        
        \multicolumn{2}{|c|}{\multirow{2}{*}{\textbf{Features}}} & \multicolumn{2}{c|}{\textbf{PETA}}  &
        \multicolumn{2}{c|}{\textbf{PA-100K}} & \multicolumn{2}{c|}{\textbf{RAP}} \\
        & & Correct & All  & Correct & All & Correct & All
        \\\hline\hline
        
        \multirow{3}{*}{Subject Pose}  
        & Frontal       & 0.387 & 0.383 
                        & 0.363  & 0.355 
                        & 0.350 & 0.348 \\
        
        & Sideways      & 0.223 & 0.250  
                        & 0.356  & 0.368 
                        & 0.309 & 0.315 \\
                        
        & Backside      & 0.390 & 0.366  
                        & 0.282  & 0.277  
                        & 0.341 & 0.337 \\
         \hline
        \multirow{3}{*}{KP Confidence}
        & Face          & 0.823 $\pm$  0.105 & 0.821 $\pm$  0.107 
                        & 0.820 $\pm$  0.124  & 0.813 $\pm$  0.128   
                        & 0.839 $\pm$  0.117 & 0.835 $\pm$  0.125 \\
        
        & Upper Body    & 0.781 $\pm$  0.109 & 0.775 $\pm$  0.114  
                        & 0.773 $\pm$  0.099  & 0.765 $\pm$  0.104  
                        & 0.820 $\pm$  0.094 & 0.817 $\pm$  0.098 \\
        
        & Lower Body    & 0.764 $\pm$  0.108 & 0.759 $\pm$  0.112  
                        & 0.768 $\pm$  0.087 & 0.761 $\pm$  0.093     
                        & 0.786 $\pm$  0.115 & 0.783 $\pm$  0.118 \\
        
        \hline
        \multirow{3}{*}{Image} 
        & Resolution    & 0.211 $\pm$  0.131 & 0.207 $\pm$  0.132
                        & 0.066 $\pm$  0.120  & 0.054 $\pm$  0.070  
                        & 0.223 $\pm$  0.157 & 0.227 $\pm$  0.155 \\
        
        & Luminosity    & 0.417 $\pm$  0.116 & 0.415 $\pm$  0.119 
                        & 0.452 $\pm$  0.160 & 0.462 $\pm$  0.158 
                        & 0.478 $\pm$  0.152  & 0.475 $\pm$  0.153   \\
        
        & Blurriness    & 0.148 $\pm$  0.188 & 0.143 $\pm$  0.175 
                        & 0.114 $\pm$  0.120 & 0.110 $\pm$  0.118     
                        & 0.115 $\pm$  0.070 & 0.114 $\pm$  0.070  \\

        \hline
    \end{tabular}
    \label{tabel:method_prediction}
\end{table*}

The results suggest no significant differences between images where gender is correctly inferred and all test set images, for all the evaluated features. This experiment points out that a general feature evaluation of the correctly gender-inferred images might not be enough to identify how each feature contributes to gender inference accuracy.

\subsection{Interpreting Gender Inference}
\label{sec:shap_results}




With the intent to interpret feature values of different quality images and their relation to gender inference, we grouped all models' predictions, for all datasets, and analyze their correlation with gender inference, as described in Section \ref{method:shapley}. 



\begin{figure}[!tb]
\centering
\includegraphics[width=3.5in]{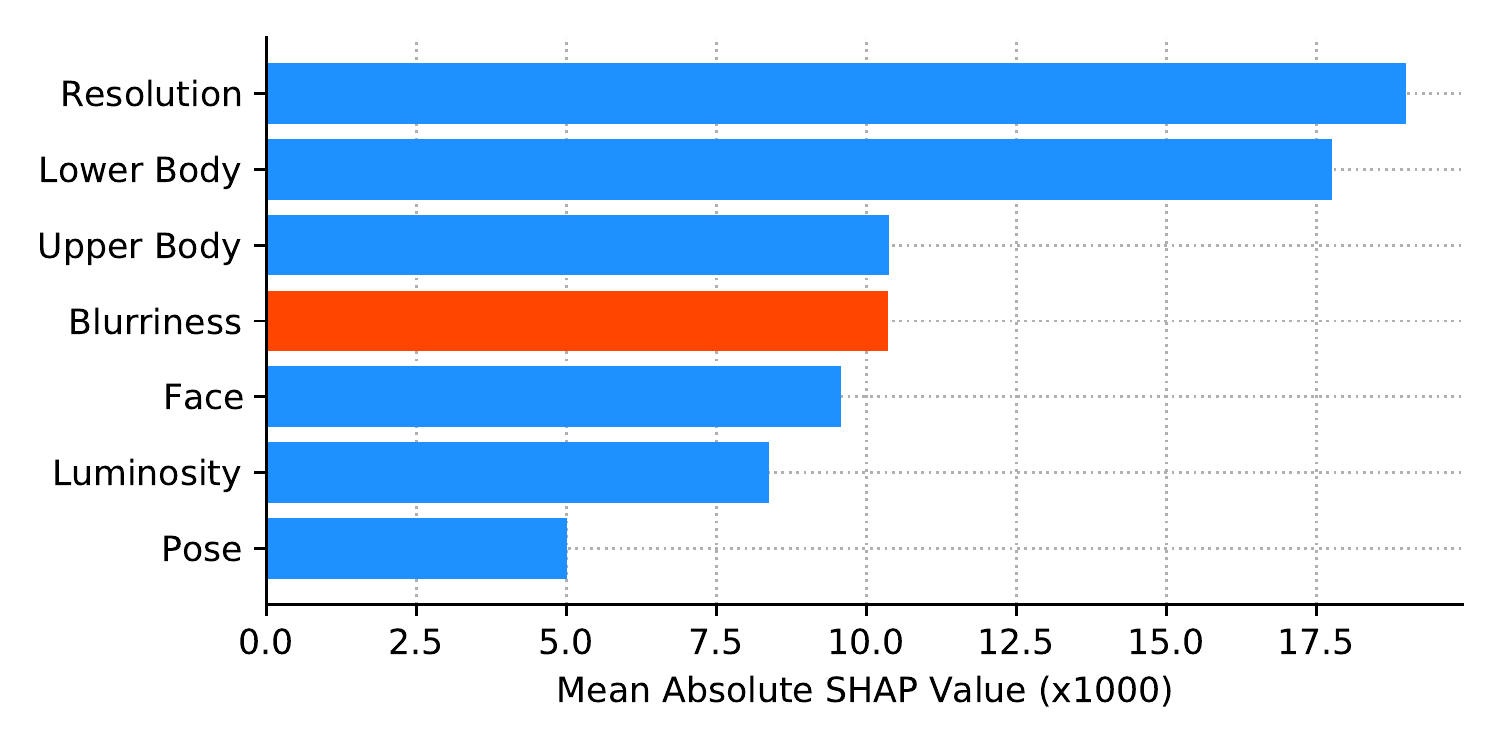}
\caption{Bar plot of the mean absolute SHAP values for image and subject-based features. Blue colored bars correlate with gender correct inference, while the red one correlates with gender misinference. Features are ordered by its influence in gender inference accuracy.}
\label{fig:mean_shap_value_all}
\end{figure}

By analyzing the bar plot for each feature mean absolute SHAP values, in Fig. \ref{fig:mean_shap_value_all}, we conclude that image resolution and lower body KP confidence are the most important characteristics. Bars in blue are positively correlated values, while the red bar denotes a negatively correlated feature. In this case, high values of blurriness contribute to gender misinference. The resolution importance was expected \emph{a priori}, given the difficulties in discriminating gender in poor resolution data. The lower body importance might be related to hip or footwear influence, which is likely to occur when the face is fully occluded, forcing models to analyze different parts for gender inference. Subjects' pose is the least influential factor among those evaluated.

\begin{figure*}[!ht]



\subfloat[Face]{\includegraphics[width=2.2in]{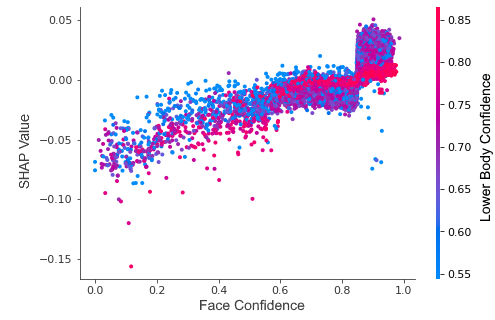}}
\quad
\subfloat[Lower Body]{\includegraphics[width=2.2in]{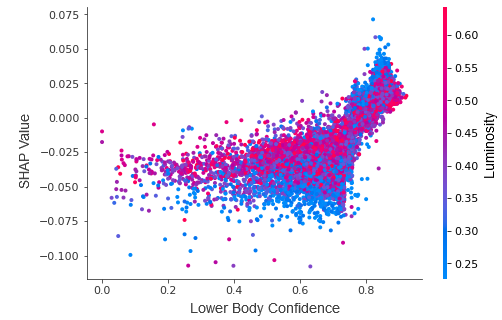}}
\quad
\subfloat[Upper Body]{\includegraphics[width=2.2in]{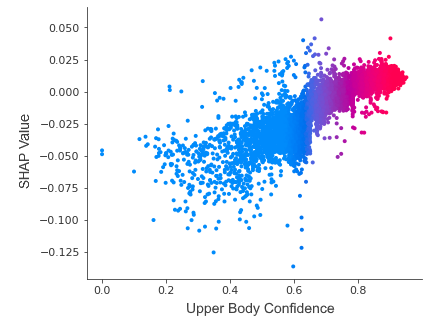}}

\subfloat[Resolution]{\includegraphics[width=2.2in]{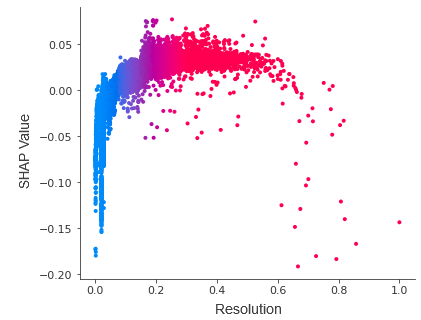}}
\quad
\subfloat[Luminosity]{\includegraphics[width=2.2in]{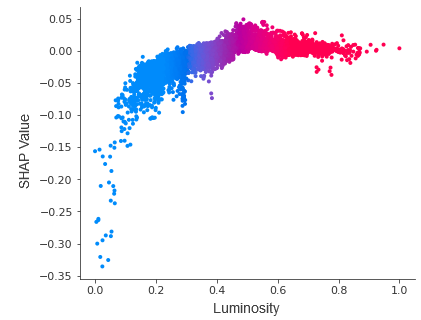}}
\quad
\subfloat[Blurriness]{\includegraphics[width=2.2in]{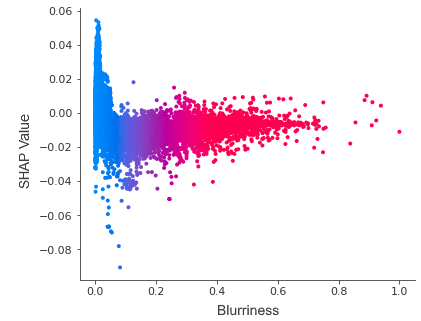}}

\caption{Dependence scatter plots of Face (a), Lower Body (b), Upper Body (c), Resolution (d), Luminosity (e), and Blurriness (f) SHAP values. Face and Lower Body plots present a correlation with the named feature and Lower Body and Luminosity, respectively. These plots contain a right $y$ axis with color variation, representing correlated feature value variance. Color variation from blue to red in the other plots represent increasing values of the feature. Image-based features are normalized. 
}
\label{shap_all}
\end{figure*}

To better understand each feature's importance in gender inference, we evaluate the dependence plots for some features. Fig. \ref{shap_all} shows the results for the face, lower, and upper body KP confidence (subjects-based factors) and resolution, luminosity, and blurriness dependence plots (image-based features). Negative SHAP values are linked to gender misinference, while positive ones are linked to correct inference. The face and lower body KP confidence plots correlate each corresponding feature with lower body and luminosity, respectively. For the plots with no feature correlation, the color variation from blue to red represents an increase in feature values. We conclude that higher KP confidence values translate into gender correct inference. For lower and upper body, KP confidence values above 0.7 are linked to correct inference, while face requires KP confidence values above 0.8 for the same effect. Low values of resolution heavily negatively impact gender inference and are positively influential in higher values. Luminosity only influences gender inference when it has low values and tends to have a neutral role with value increase. Blurriness has a positive influence on low values and a negative influence when its value arises. An increase in blurriness does not translate into more incorrect gender inference.

Regarding the feature correlation plots, for face confidence values lower than 0.8, higher confidence of lower body tends to translate into higher SHAP values. This indicates that models tend to focus on the lower body portion to perform inference when the face is not visible. Regarding the lower body, increased luminosity generally improves the accuracy, which is more evident when KP confidence is between 0.4 and 0.7. This suggests that when a subject's lower body is in uncooperative poses or minor occluded, better illumination might dissipate the difficulty in interpreting such poses for gender models.

The observed results suggest that higher resolution images with good illumination, low blurriness, and cooperative poses (translated in high KP confidence values) are important factors for correct gender inference. Furthermore, the face might not be reliable for \textit{in-the-wild} gender inference since it requires higher confidence values than those of the lower and upper body for accurate inference.

\begin{figure}[!tb]
\centering
\includegraphics[width=3in]{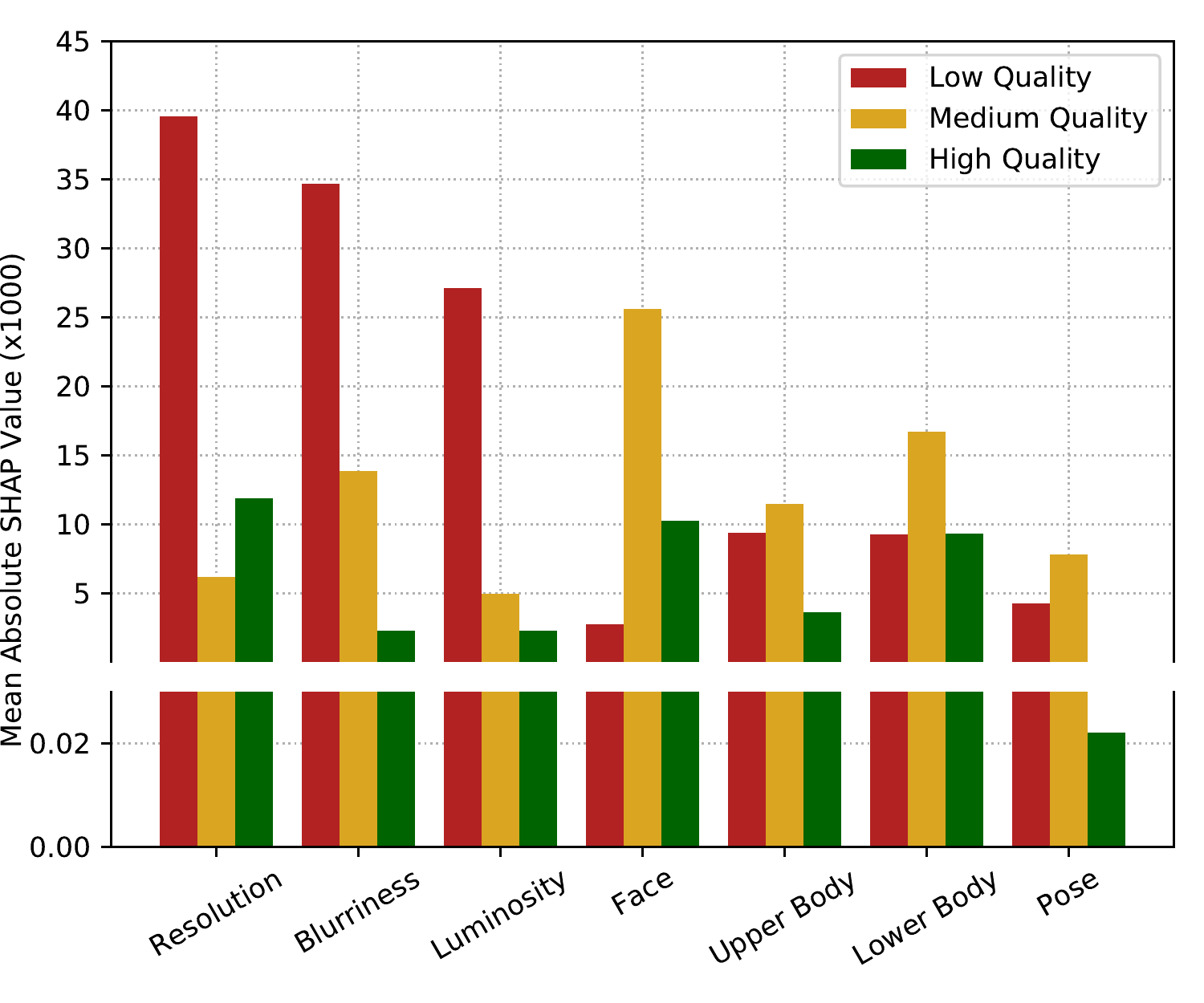}
\caption{Bar plots of the evaluated features, presenting the SHAP values for low, medium, and high quality image datasets. Datasets are represented by different colors. High SHAP values represent high feature relevance for gender correct inference. The bar plot is broken in the vertical axis for better visualization.}
\label{fig:shap_all_quality}
\end{figure}

In addition to the overall analysis of features influence, we further investigate if image quality can change the importance of each image/subject-based feature. For this, we analyze feature importance in each dataset, translated in low (PETA), medium (PA-100K), and high (RAP) quality images. We present the bar plots of our analysis in Fig. \ref{fig:shap_all_quality}. All SHAP values refer to the mean absolute values of each feature.

We can conclude that image-based features are highly influential in low quality images, given the high SHAP values they present in these conditions. This analysis also suggests that subject-based features are not reliable for gender inference in scenarios where image quality is low. If we examine image-based features across image quality variance, we conclude that they tend to have less importance as quality increases. Subject-based features increase their importance from low to medium quality images. Additionally, in this quality range, subject-based features are more important than image-based, illustrating a shift of feature importance for gender inference. In high quality images, the SHAP values are all lower than in other quality range images. This indicates that each feature loses importance in high quality data, where the models can analyze different features reliably. The pose is the least influential feature, presenting an insignificant influence in high quality images.


These findings can be applied to improve model inference. Our results suggest that, if the inference is done using low quality images, it is advised to collect new data with improved image-based features. However, assuming medium quality data usage, image quality increase (such as having better illumination or resolution) is not as important as improving subject-based features. In high quality images, gender inference accuracy improvements might not be achieved through data changes, given the relatively low influence of the analyzed features for gender inference. Redefining models' architecture might be a better strategy to improve overall accuracy.

\subsection{Face Importance in Gender Inference}

The results described in section \ref{sec:shap_results} suggest that face importance for gender inference might be directly related to image quality. To confirm this idea, we analyze the reliability of face-based gender inference given different data quality.  We start by training our face-based gender model in Adience, one of the most challenging face gender datasets. We present the accuracy obtained for our model in Table \ref{table:adience}, in comparison to results reported in the literature (5-fold cross-validation \cite{eidinger2014age} values).

\begin{table}[!tb]
    \centering
    \caption{Comparison of the proposed face-based gender model, in the Adience dataset.}
    \begin{tabular}{|c|*{6}{c|}}\hline
        \textbf{Methods}
        &\makebox[4em]{\textbf{Acc (\%)}}\\\hline\hline
        Lee \emph{et al.} (2018) \cite{lee2018joint} & 85.16 \\
        Levi and Hassner (2015) \cite{levi2015age} & 86.80 \\
        Hung \emph{et al.} (2019) \cite{hung2019increasingly} & 89.50 \\
        Zhang \emph{et al.} (2017) \cite{zhang2017age} & 92.43 \\\hline
        Face-Model & 93.79 \\\hline
    \end{tabular}
    \label{table:adience}
\end{table}

The obtained accuracy is better than the reported ones, indicating that our model represents a good face-based model. To assess its applicability in wild conditions, we evaluate its \textit{cross-domain} performance using  Adience, PETA, PA-100K, and RAP datasets. To keep the Adience values' comparison as fair as possible, we train and evaluate the models in PETA, PA-100K, and RAP using only face images of frontal subjects of each dataset.  We report the results in Table \ref{table:cross-eval-face}. The Adience result is based on the proposed face-based gender model when training and testing on the first folder of the 5-fold cross-validation. 

We can observe that the Adience-trained model performs poorly in other datasets, illustrating that face-based training is not adequate for \textit{in-the-wild} inference. However, the dataset where this model performs the best is RAP, characterized as a high quality images set. Similarly, RAP-trained models achieve higher accuracy performance in Adience than in PETA or PA-100K. These results indicate that Adience and RAP face dataset have similarities, which corroborates that face importance is closely related to increased image quality. To further explore this concept, we assess image quality and the subject's pose effect on face importance. For this, we train our face-based gender model in all dataset images (full-body) and evaluate its performance in face images of the correspondent dataset. The procedure is repeated when training the model only in frontal images. We present the obtained results in Table \ref{table:face_imp}.

\begin{table}[!tb]
    \centering
    \caption{Gender $mA$ of face-based gender model, trained and evaluated in within and cross-domain settings. 
    Only frontal face images of PETA, PA-100K, and RAP datasets were considered.
    }
    \begin{tabular}{|c|*{9}{c|}}\hline
        \backslashbox[20mm]{\textbf{Train}}{\textbf{Eval}} & \makebox[3em]{Adience}
        & \makebox[3em]{PETA} 
        & \makebox[3em]{PA-100K} 
        & \makebox[3em]{RAP} 
        \\\hline\hline
        
        Adience &
        
         94.83 &
        
        
        65.56 &
        
        55.55 &
        
        77.85 \\ 
        
        PETA &
        
        64.96 &
        
        
         91.16 &
        
        74.60 &
        
        82.43 \\
        
        PA-100K &
        
        75.62 &
        
        
        72.52  &
        
         84.71 &
         
        79.72  \\
        
        RAP &
        
        75.49 &
        
        
        68.09 &
        
        68.32 &
        
         95.31 \\
        
        \hline
    \end{tabular}
    \label{table:cross-eval-face}
\end{table}

\begin{table}[!tb]
    \centering
    \caption{Gender $mA$ of face-based gender model, trained for \textit{Any} and \textit{Frontal} pose images, and evaluated in frontal image faces (\textit{Face}) and full-body images of corresponding pose (\textit{Body}). \textit{Any Pose} refers to all dataset images and \textit{Frontal Pose} images refer to frontal images. 
    \texttt{FI} stands for Face Importance.}
    \begin{tabular}{|c|*{6}{c|}}\hline
        \multicolumn{1}{|c|}{} & \multicolumn{3}{c|}{\textbf{Any Pose}} & \multicolumn{3}{c|}{\textbf{Frontal Pose}}  \\\hline\hline
        \textbf{Dataset} & 
        \multicolumn{1}{c|}{Face} & 
        \multicolumn{1}{c|}{Body} &
        \multicolumn{1}{c|}{FI (\%)} &
        \multicolumn{1}{c|}{Face} & 
        \multicolumn{1}{c|}{Body} &
        \multicolumn{1}{c|}{FI (\%)}\\\hline\hline
        PETA & 
        57.60 & 93.24 & 17.58 &
        61.09 & 92.52 & 26.08 \\

        PA-100K &
        52.52 & 91.06 & 6.14 &
        57.03 & 91.86 & 16.79 \\
        
        RAP & 
        73.63 & 96.09 & 51.27 &
        75.31 & 95.98 & 55.05 \\
        
        \hline
    \end{tabular}
    \label{table:face_imp}
\end{table}

The results observed support that higher quality images directly influence the importance of face in gender inference. Furthermore, when we train the model with frontal data and evaluate using only facial images, the face importance increased for all three datasets, suggesting that the face importance factor is also linked to its availability. This finding suggests an implicit attention towards the face, motivated solely by guaranteed face access, via training in frontal images only.

\section{Conclusions}
\label{sec:conclusion}

Soft biometrics inference in wild conditions is hindered by factors such as partial occlusions, subjects' pose, and varying levels of image quality. Considering, in particular, the gender trait, this work analyzes the image and subject-based features that are the most/least linked to correct inference of state-of-the-art models. In our analysis, we assess the combined gender inference of five state-of-the-art PAR models in well known PAR datasets. We evaluate the effectiveness of face-based gender inference models in surveillance scenarios, also considering the effect of pose. Our results suggest that image-based features are more important in low-quality images, while the importance of subject-based features is linked to an increase in image quality. Moreover, the relevance of the facial region to correctly infer gender is directly related to high-quality data. Finally, we show an implicit face attention approach, linked to its guaranteed availability, via frontal image training.

Our findings provide the basis for designing more robust gender inference models, particularly suited to work \textit{in-the-wild}, where image and subject-based features highly vary. The improvement of face-based gender inference via an implicit face attention mechanism paves the way to incorporate similar approaches in future works.


%



\section*{Acknowledgment}

This work is funded by FCT/MEC through national funds and co-funded by FEDER - PT2020 partnership agreement under the project UIDB//50008/2020. 
Also, it was supported by operation Centro-01-0145-FEDER-000019 - C4 - Centro de Compet\^{e}ncias em Cloud Computing, co-funded by the European Regional Development Fund (ERDF) through the Programa Operacional Regional do Centro (Centro 2020), in the scope of the Sistema de Apoio \`{a} Investiga\c{c}\~{a}o Cient\'{i}fica e Tecnol\'{o}gica - Programas Integrados de IC\&DT, and supported by FCT - Fundação para a Ciência e Tecnologia through the research grant 2020.09847.BD.


\ifCLASSOPTIONcaptionsoff
  \newpage
\fi

\end{document}